\documentclass[conference, final]{IEEEtran}
\IEEEoverridecommandlockouts
\usepackage{cite}
\usepackage{amsmath,amssymb,amsfonts}
\usepackage{algorithmic}
\usepackage{graphicx}
\usepackage{textcomp}

\usepackage[inline]{enumitem}
\usepackage{algorithmic}
\usepackage{graphicx}
\usepackage{textcomp}
\usepackage{xcolor}

\usepackage{witharrows}

\usepackage[flushleft]{threeparttable}
\usepackage{tablefootnote}
\usepackage{array}

\newcolumntype{P}[1]{>{\centering\arraybackslash}m{#1}}
\usepackage{multicol}
\usepackage{multirow}
\usepackage{tabulary}
\usepackage{colortbl}
\definecolor{mygray}{gray}{0.90}
\usepackage{makecell}
\usepackage{booktabs}
\usepackage{subcaption}
\usepackage{stmaryrd}
\usepackage{float}
\usepackage{pifont}

\usepackage{arydshln}
\colorlet{mygray}{gray!15!white}

\usepackage[switch,columnwise]{lineno}

\usepackage{url,hyperref,microtype}
\hypersetup{
    colorlinks=true,
    citecolor=blue,
    linkcolor=blue,
    filecolor=magenta,      
    urlcolor=black,
}

\def\BibTeX{{\rm B\kern-.05em{\sc i\kern-.025em b}\kern-.08em
    T\kern-.1667em\lower.7ex\hbox{E}\kern-.125emX}}
\begin{document}

\title{FUSE: Frame-Unified Stress Estimation from Facial Video\\
}

\author{

\IEEEauthorblockN{Stefanos Gkikas}
\IEEEauthorblockA{\textit{Honda Research Institute Japan} \\
Wako City, Japan \\
stefanos.gkikas@jp.honda-ri.com}

\and

\IEEEauthorblockN{Thomas Kassiotis}
\IEEEauthorblockA{\textit{Department of Electronic Engineering} \\
\textit{Hellenic Mediterranean University}\\
Chania, Greece \\
ddk305@edu.hmu.gr}

\and

\IEEEauthorblockN{Yang Guo}
\IEEEauthorblockA{\textit{Faculty of Information Science}
\textit{and Engineering} \\
\textit{Ocean University of China}\\
Qingdao, China \\
diw85827@gmail.com}

\and

\IEEEauthorblockN{Guangliang Li}
\IEEEauthorblockA{\textit{Faculty of Information Science}
\textit{and Engineering} \\
\textit{Ocean University of China}\\
Qingdao, China \\
guangliangli@ouc.edu.cn}

\and

\IEEEauthorblockN{Giorgos Giannakakis}
\IEEEauthorblockA{\textit{Department of Electronic Engineering} \\
\textit{Hellenic Mediterranean University}\\
Chania, Greece \\
ggian@hmu.gr}

}

\maketitle

\begin{abstract}
Automatic stress detection from facial video offers a practical path to non-intrusive affect monitoring, yet existing video-based approaches commonly decompose full recordings into short temporal windows before classification. This design introduces additional choices regarding window length, overlap, and aggregation, while limiting direct analysis of temporal information across the entire recording. In this study, we present FUSE (\textbf{F}rame-\textbf{U}nified \textbf{S}tress \textbf{E}stimation), a facial-video stress detection framework that processes complete recordings as a single input without temporal windowing or external segmentation. The name reflects the defining operation of the method: rather than dividing a recording into short clips, all frames are fused into one unified two-dimensional representation from which the stress state is estimated. This unification is realized by folding the temporal dimension into the channel dimension of the spatial representation, and the resulting high-dimensional input is processed using a unified asymmetric-attention architecture. At a temporal stride of $\tau=1$, FUSE retains the full $120$-second recording as one input, corresponding to $3{,}600$ frames at $30$~fps. Experiments on a $58$-subject stress dataset using a stratified subject-level protocol evaluate seven temporal-stride configurations, ranging from full-frame input to sparse subsampling. FUSE achieves the highest test accuracy of $69.44\%$ at $\tau=15$, while the full-frame configuration remains competitive at $69.03\%$. Across the stride range, computational cost varies from $12.48$ to $348.78$ GFLOPs, showing the trade-off between temporal density and efficiency. These results demonstrate that temporal windowing is not required for effective facial-video stress detection in this setting, and that complete-recording inference can be achieved within a single unified architecture.

\end{abstract}



\section{Introduction}

Stress arises as a coordinated physiological and psychological response to perceived demands, engaging the autonomic nervous system and triggering neuroendocrine cascades that vary in intensity and duration \cite{goldstein_2023, hellhammer_wust_2009}. Its manifestations range from brief, situationally bounded episodes to prolonged chronic states, each associated with distinct physiological signatures and long-term health consequences. Questionnaire-based instruments such as the Perceived Stress Scale are widely used in clinical and research settings to quantify subjective stress burden \cite{cohen_kamarck_1983}, but retrospective self-reports are susceptible to recall bias and poorly suited to capturing fine-grained temporal variation in stress levels \cite{shiffman_stone_2008}. Salivary cortisol serves as a validated neuroendocrine index of the stress response, yet the invasive nature of sample collection and the delayed temporal dynamics of cortisol secretion limit its applicability for continuous, real-time monitoring \cite{hellhammer_wust_2009}.

The scale of stress as a public health problem has grown substantially over recent decades. A large-scale analysis of nationally representative survey data spanning $146$ countries found that reported stress levels roughly doubled over an $18$-year period, with disparities widening across demographic and socioeconomic groups \cite{canaletti_lun_2026}. In occupational contexts, psychosocial work-related exposures have been estimated to account for a measurable share of cardiovascular disease and depression cases across European countries \cite{sultantaib_villeneuve_2022}. At the biological level, chronic psychological stress has been linked to disruption of immune regulation, elevated cardiovascular risk, and depressive disorder, underscoring the broader clinical relevance of reliable stress monitoring \cite{cohen_janicki_2007}.

Reliable stress monitoring is most valuable precisely in the contexts where existing assessment methods are least practical. In clinical, occupational, and operational settings, self-report is often delayed, incomplete, or systematically biased by social desirability and demand characteristics. Contact-based wearable devices provide a continuous physiological alternative,
but face persistent challenges around user compliance, motion-induced signal degradation, and scalability to diverse deployment environments \cite{hosseini_gottumukkala_2026}. Field deployments of such systems have consistently identified data quality and signal integrity as limiting factors that constrain generalization beyond laboratory conditions \cite{neigel_vargo_2025}. These considerations motivate the development of passive, non-intrusive automated systems capable of objectively recognizing stress from signals naturally available in everyday environments \cite{giannakakis_grigoriadis_2022}.

Among non-intrusive modalities, facial video recorded by standard cameras offers a particularly practical channel for automated stress monitoring, requiring no physical contact, specialized hardware, or user compliance beyond proximity to a camera. Stress-related changes in facial behavior, including eye activity, mouth movements, and head dynamics, carry discriminative information for separating stressed from neutral and relaxed states \cite{giannakakis_pediaditis_2017}, and Facial Action Unit representations provide a structured encoding of these signals for automatic recognition \cite{giannakakis_koujan_2020}. Deep learning has substantially advanced the state of the art in video-based stress recognition, with spatiotemporal architectures and AU-based pipelines achieving consistent improvements over handcrafted feature approaches \cite{kyrou_kompatsiaris_2025}. A shared limitation of existing methods, however, is their reliance on fixed-length temporal windows or short clips as the unit of analysis: full recordings are partitioned into segments of a few seconds before classification, introducing additional design choices and discarding the temporal context that spans the window
boundaries \cite{zhang_feng_2020, jeon_bae_2021, valergaki_nicodemou_2026}.

In this work, we present FUSE, a framework for automatic stress detection in facial video that processes entire recordings as a single input, without any temporal windowing or external segmentation. 
The full video sequence is encoded through axis folding and processed by a unified asymmetric attention model, whose internal spatial token segmentation handles the resulting high-dimensional representation without imposing any temporal partitioning on the input. Experiments on a $58$-subject stress dataset, evaluated under a stratified subject-level protocol, assess performance across temporal stride configurations ranging from dense sampling to full-frame input, demonstrating that the architecture natively accommodates complete recordings without modification.
Automatic human-state recognition has been investigated across a range of signals, including stress and pain estimation from electrodermal activity and other biosignals \cite{gkikas_eda_stress_prai_2026, gkikas_kyprakis_eda_2025, gkikas_tiny_2025, gkikas_kyprakis_resp_2025}, emotion recognition from EEG \cite{gkikas_guo_eeg_prai_2026}, and cognitive workload assessment \cite{gkikas_workload_acii_2026}, complementing the facial-video stress detection addressed here.

\section{Related Work}
\label{related_work}
Video-based approaches to stress recognition have been explored as a contact-free alternative to physiological monitoring, leveraging the behavioral information encoded in facial dynamics. Initial methods extracted handcrafted features from facial regions, including eye-related activity, mouth movements, head-motion statistics, and camera-based heart rate estimates, demonstrating their discriminative power in distinguishing stressed from neutral and anxious states \cite{giannakakis_pediaditis_2017}. Deep learning methods replaced handcrafted pipelines by jointly learning face- and action-level representations from video, achieving improved accuracy over feature-engineering baselines on a purpose-built video stress dataset \cite{zhang_feng_2020}. Spatiotemporal architectures extended this direction by modeling both spatial regions and temporal changes in facial appearance, using fixed-length clips as the input unit for learning stress-related facial
dynamics \cite{jeon_bae_2021}. Beyond laboratory-induced stress protocols, recent work has investigated facial video-based stress detection under naturalistic conditions, collecting data without artificial contextual constraints to improve ecological validity \cite{ding_xu_2025}.

Facial Action Units provide a physiologically grounded encoding of facial muscle activity and have been adopted as an interpretable intermediate representation for automatic stress recognition from video. AU-based classifiers have been applied to distinguish stressed from neutral states in a contact-free setting, with specific action unit patterns consistently emerging as discriminative across subjects and stressor types \cite{giannakakis_koujan_2020}. Deep learning pipelines for automatic AU recognition have been adapted to the stress domain, combining geometric deformation features with deep appearance descriptors extracted from facial video to classify affective states \cite{giannakakis_koujan_2022}. Explainable AI methods have been integrated into AU-based stress recognition to identify the facial muscle activations that are most predictive of stress, thereby providing interpretable evidence alongside model outputs \cite{giannakakis_roussos_2025}. 
An explainable graph attention network operating on differential facial action units has been proposed for stress recognition \cite{kassiotis_stressgat_acii_2026}.
Comprehensive surveys of deep learning for stress detection identify facial video analysis as a prominent research direction, with video-based approaches showing consistent improvements alongside advances in spatiotemporal representation learning \cite{kyrou_kompatsiaris_2025}.

A common design choice in video-based stress recognition is to convert continuous recordings into predefined temporal units before classification. For example, Zhang et al. \cite{zhang_feng_2020} partitioned each 2-min recording into 15-s samples, while Jeon et al. \cite{jeon_bae_2021} modeled stress from short 2-s facial clips. More recent work has followed a similar temporal-reduction strategy, including non-overlapping windows for Transformer-based stress estimation \cite{valergaki_nicodemou_2026}, or frame-level facial feature extraction followed by temporal and
frequency-domain aggregation \cite{ding_xu_2025}. Although these strategies make learning computationally tractable, they introduce additional design choices, such as clip duration, window length, overlap, and aggregation strategy. These choices are dataset- and context-dependent and may limit the model's ability to exploit temporal information that spans predefined boundaries. Recent surveys confirm the growing role of deep learning in stress detection across facial, behavioral, physiological, and multimodal signals \cite{kyrou_kompatsiaris_2025}, but full-recording video analysis without external temporal windowing remains underexplored. 

Efficient long-video modeling has been widely studied through spatiotemporal token designs, such as the tubelet embedding of ViViT \cite{arnab_dehghani_2021}, which nonetheless retain an explicit temporal axis. FUSE instead adapts the asymmetric-attention principle of the Perceiver \cite{jaegle_gimeno_2021}, folding time into the channel dimension so that the full recording is processed as a single spatial representation rather than a growing spatiotemporal token set. Transformer-based and modality-agnostic architectures have similarly been employed for affective assessment from facial video and physiological signals \cite{gkikas_tsiknakis_tnt_2023, gkikas_tachos_2024, gkikas_tsiknakis_painvit_2024, gkikas_rojas_painformer_2025}.


\section{Methodology}
\label{sec:methodology}

\subsection{Video Tokenization}

The proposed framework encodes facial video input into a token representation without temporal segmentation or modality-specific components, enabling a single model to process complete recordings of varying length. For a video of $T$ frames
captured at $30$~fps and subsampled at temporal stride $\tau$, the retained sequence contains $L = \lfloor T / \tau \rfloor$ frames, each a $224\times224$ RGB image. At a stride $\tau=1$ applied to a $120$-second recording, this yields $L=3{,}600$ frames, corresponding to the complete recording without any external windowing or temporal segmentation. The temporal dimension is folded into the channel dimension---a step referred to as \textit{axis folding}---creating a tensor of shape $H\times W\times 3L$. This preserves the spatial structure of each frame while packing the entire temporal sequence into a single 2D representation:
\begin{equation}
\mathbf{X} \in \mathbb{R}^{B \times H \times W \times 3L},
\end{equation}
where $B$ denotes the batch size and $H=W=224$.
Geometric information is incorporated by encoding each spatial position $\mathbf{p}\in[-1,1]^2$ using Fourier features. The model uses $K=6$ frequency bands and a maximum frequency $f_{\max}=10$. Since the input has $D=2$ spatial axes, the Fourier encoding adds $D(2K+1)=26$ positional features. The encoding is:
\begin{equation}
\begin{split}
\gamma(\mathbf{p}) = \bigl[&\sin(\pi s_1 \mathbf{p}),\ \cos(\pi s_1
\mathbf{p}),\ \ldots,\\
&\sin(\pi s_K \mathbf{p}),\ \cos(\pi s_K \mathbf{p}),\ \mathbf{p}\bigr],
\end{split}
\end{equation}
where $\{s_k\}_{k=1}^{K}$ spans $[1, f_{\max}/2]$. The spatial axes are flattened into a sequence of $N = H\times W = 50176$ tokens, with data channels and positional features concatenated per token to form the token matrix:
\begin{equation}
\mathbf{T} \in \mathbb{R}^{B \times N \times C'},\quad C' = 3L + D(2K+1),
\end{equation}
where $D=2$ denotes the number of spatial axes. Since $D=2$ and $K=6$, the token dimension becomes:
\begin{equation}
C' = 3L + 26.
\end{equation}
The token sequence is partitioned into $S=4$ contiguous spatial groups of length $n_s = N/S = 12544$ tokens. These spatial token groups, referred to as segments throughout, divide the folded 2D representation along the token axis and carry no correspondence to temporal windows of the input video. The tokens of spatial segment $s$ are denoted $\tilde{\mathbf{T}}_s \in
\mathbb{R}^{B\times n_s\times C'}$.

\subsection{Asymmetric Attention}
\label{sec:asymmetric}

The model processes the segmented token sequence through four layers, each comprising a cross-attention block followed by $R_\ell$ self-attention blocks. A single latent state is associated with each spatial segment, instantiated at runtime by replicating a shared initialization vector derived from a set of $M_0=32$ learnable global parameters
$\{\boldsymbol{\ell}_m\}_{m=1}^{M_0}$:
\begin{equation}
\boldsymbol{\ell}_{\mathrm{init}} = \frac{1}{M_0}\sum_{m=1}^{M_0}
\boldsymbol{\ell}_m \in \mathbb{R}^{d_0},
\end{equation}
where $d_0=128$. These segment states are not independently learnable; segment-specific representations emerge through the attention updates.

\textbf{Cross-attention.} At each layer $\ell$, each segment state aggregates information exclusively from its corresponding token subset through cross-attention:
\begin{equation}
\mathbf{e}_s^{(\ell)} = \mathbf{e}_s^{(\ell-1)} +
\mathrm{Attn}\bigl(\mathbf{e}_s^{(\ell-1)},\ \tilde{\mathbf{T}}_s\bigr),
\end{equation}
where $\mathbf{e}_s^{(\ell-1)}\in\mathbb{R}^{B\times 1\times d_\ell}$
provides the queries and $\tilde{\mathbf{T}}_s\in\mathbb{R}^{B\times
n_s\times C'}$ provides the keys and values. This operation is asymmetric: the query side consists of a single vector of dimension $d_\ell$, while the key-value side spans $n_s \gg 1$ token vectors of dimension $C'$. The resulting attention matrix is $1\times n_s$, not square, and the query and key-value spaces differ in both size and dimensionality. All $S$ segments are processed in parallel by packing into the batch dimension, without altering the underlying computation. Cross-attention uses a single head at all layers, with per-layer head dimensions of $\{64, 48, 32, 16\}$.

\textbf{Self-attention.} After cross-attention, all segment states are stacked to form the segment-state matrix $\mathbf{E}^{(\ell)} \in \mathbb{R}^{B\times S\times d_\ell}$. Self-attention is applied across all $S$ segment states, enabling global information exchange:
\begin{equation}
\mathbf{E}^{(\ell)} \leftarrow \mathbf{E}^{(\ell)} +
\mathrm{Attn}\bigl(\mathbf{E}^{(\ell)},\ \mathbf{E}^{(\ell)}\bigr),
\end{equation}
repeated $R_\ell \in \{8, 6, 4, 2\}$ times per layer for $\ell = 0,\ldots,3$. Self-attention uses multi-head attention with per-layer head counts of $\{8, 6, 4, 2\}$ and per-layer head dimensions of
$\{64, 48, 32, 16\}$.

\begin{figure}
\begin{center}
\includegraphics[scale=0.23]{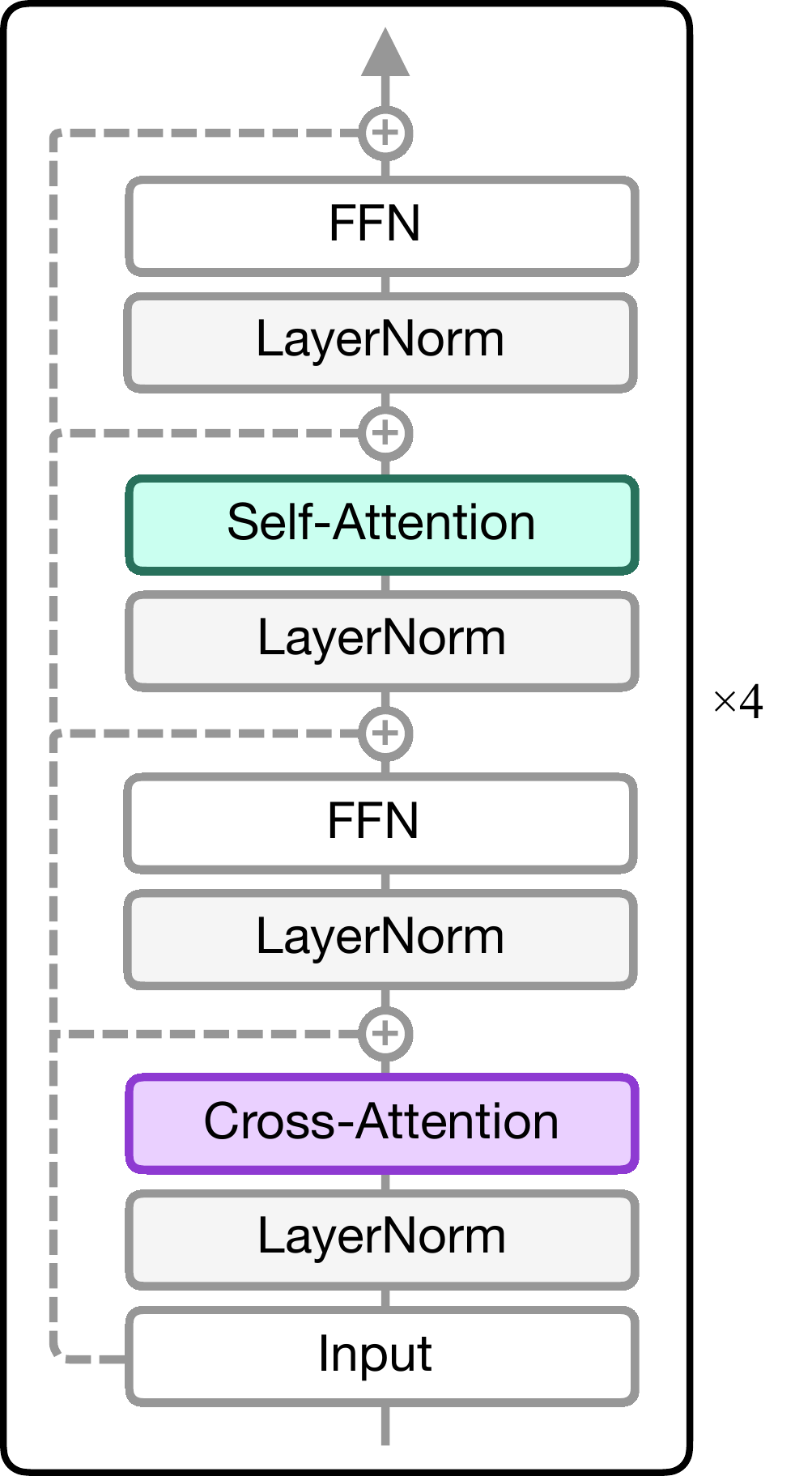}
\end{center}
\caption{FUSE layer block, repeated four times. Each layer comprises a cross-attention sub-block, in which a single latent segment state attends over its corresponding spatial token group, followed by $R_\ell$ self-attention sub-blocks that exchange information globally across all segment states. All sub-blocks use pre-layer normalization and residual connections.}
\label{model}
\end{figure}

\textbf{Hierarchical segment-state compression.} Across the four layers, the segment-state dimensionality decreases progressively as $d_\ell \in \{128, 112, 96, 80\}$. At each layer transition, the segment representations are projected to the new dimensionality through a linear transformation when required. The number of spatial segment states remains fixed at $S=4$
throughout all layers, with one state per token group. After the final layer, the final segment states $\mathbf{E}^{(4)} \in \mathbb{R}^{B\times S\times 80}$ are averaged across segments and passed through a linear classification head. Both attention operations use pre-layer normalization and residual connections, with attention and feedforward dropout
of $0.10$ applied uniformly.
The complete set of architectural hyperparameters is reported in Table~\ref{tab:architecture} and the layer structure is illustrated in Fig.~\ref{model}.

\begin{table}
\caption{Architectural hyperparameters of FUSE.}
\label{tab:architecture}
\begin{center}
\scriptsize
\begin{threeparttable}
\begin{tabular}{>{\raggedright\arraybackslash}m{5.5cm} P{2.0cm}}
\toprule
\textbf{Hyperparameter} & \textbf{Value} \\
\midrule
\midrule
Depth                                      & 4                \\\hdashline
Latent pool size ($M_0$)                   & 32               \\\hdashline
Latent dimension ($d_\ell$)                & 128, 112, 96, 80 \\\hdashline
Cross-attention heads                      & 1, 1, 1, 1       \\\hdashline
Cross-attention head dimension             & 64, 48, 32, 16   \\\hdashline
Self-attention heads                       & 8, 6, 4, 2       \\\hdashline
Self-attention head dimension              & 64, 48, 32, 16   \\\hdashline
Self-attention blocks per cross ($R_\ell$) & 8, 6, 4, 2       \\\hdashline
Spatial segments ($S$)                     & 4                \\\hdashline
Attention dropout                          & 0.10             \\\hdashline
Feedforward dropout                        & 0.10             \\\hdashline
Fourier frequency bands ($K$)              & 6                \\\hdashline
Maximum frequency ($f_{\max}$)             & 10               \\
\bottomrule
\end{tabular}
\begin{tablenotes}[para,flushleft]
\scriptsize
\item Per-layer values are listed from layer~$1$ to layer~$4$.
\end{tablenotes}
\end{threeparttable}
\end{center}
\end{table}

\section{Experimental Evaluation \& Results}
This section presents the experimental evaluation of FUSE across multiple temporal stride configurations. All experiments are evaluated under a binary classification setting. Validation performance is reported using macro-averaged accuracy, precision, and F1 score. Test performance is reported using macro-averaged accuracy.

\subsection{Dataset and Protocol}
\label{ssec:data_collection}
A stress dataset comprising $58$ adults ($24$ men, $34$ women) aged $26.9\pm4.8$ years was used in this study. The experimental protocol comprised four stress-induction phases: social exposure, emotional recall, mental workload, and stressful video stimuli. Each participant completed $11$ tasks in total, comprising $4$ neutral, $6$ stress-inducing, and $1$ relaxation task, as detailed in Table~\ref{tbl:tasks}. The social exposure phase included a psychologist-led interview emphasizing negative personality traits. The emotional recall phase required participants to relive a past stressful event in real time. Mental workload was induced through a modified Stroop Color-Word Test \cite{stroop_1935} and the Paced Auditory Serial
Addition Test \cite{tombaugh_2006}. The stressful stimuli phase presented videos depicting accidents and acrophobia; a relaxing video served as a physiological recovery baseline between induction phases to minimize carryover effects and was excluded from classification.
Stress induction was verified through heart rate monitoring, which showed a statistically significant increase during stress tasks ($p<0.05$). Subjective validation was further confirmed using Self-Assessment Manikin scales, with participants reporting significantly higher arousal and lower valence during stressful phases compared with neutral baselines.
The facial video was recorded at $60$~fps and subsampled to $30$~fps with a resolution of $608\times800$ pixels. ECG was recorded continuously on a single channel at a sampling rate of $1$~kHz. This study uses facial video as the sole input modality. Binary classification is applied to distinguish neutral from stress conditions. The study received approval from the local Research Ethics Committee (approval no.~155/12-09-2022). All participants provided informed consent. The dataset is available for non-commercial research upon request.\footnote{\url{https://github.com/ggian/stress_dataset}}
Subjects are partitioned into training, validation, and testing sets at the subject level, ensuring no participant appears in more than one set. To avoid performance inflation due to subject-difficulty imbalance, a stratified split protocol is used. Leave-one-subject-out cross-validation is conducted across all recorded modalities to estimate per-subject difficulty, providing a ranking that is not biased toward any single signal source. Subjects are ranked by the combined z-score and assigned to four quartiles. The final split comprises $38$ training, $8$ validation, and $12$ testing subjects, with each set containing subjects from all four groups in proportion. The exact subject-level partition is reported in Table~\ref{tab:subject_split_stress} to support reproducibility and direct comparison with future work.

\begin{table}
\caption{Experimental tasks employed in this study.}
\label{tbl:tasks}
\begin{center}
\begin{threeparttable}
\begin{tabular}{P{0.5cm} P{3.5cm} P{2.0cm} P{1.0cm}}
\toprule
\# & Task & Duration (sec) & State \\
\midrule
\midrule
\multicolumn{4}{l}{\textit{Social Exposure}} \\
1  & Neutral reference        & 120 & N \\\hdashline
2  & Baseline description     & 120 & N \\\hdashline
3  & Interview                & 120 & S \\
\midrule
\multicolumn{4}{l}{\textit{Emotional Recall}} \\
4  & Neutral reference        & 120 & N \\\hdashline
5  & Recall stressful event   & 120 & S \\
\midrule
\multicolumn{4}{l}{\textit{Mental Workload}} \\
6  & Reading reference        & 120 & N \\\hdashline
7  & Stroop Colour-Word Test  & 120 & S \\\hdashline
8  & PASAT task               & 120 & S \\
\midrule
\multicolumn{4}{l}{\textit{Stressful Stimuli}} \\
9  & Relaxing video$^{*}$           & 120 & R \\\hdashline
10 & Adventure video          & 120 & S \\\hdashline
11 & Psychological pressure   & 120 & S \\
\bottomrule
\end{tabular}
\begin{tablenotes}[para,flushleft]
\scriptsize
\item N\,=\,neutral\quad S\,=\,stress \quad R\,=\,relaxed. *: Used as a physiological recovery baseline between induction
phases; excluded from binary classification.
\end{tablenotes}
\end{threeparttable}
\end{center}
\end{table}

\begin{table*}
\caption{Subject-level split by difficulty group. Subjects are ranked by
combined z-score and assigned to four quartile-based groups
(Q1\,=\,hardest, Q4\,=\,easiest).}
\label{tab:subject_split_stress}
\begin{center}
\begin{threeparttable}
\begin{tabular}{P{1.65cm} P{3.6cm} P{3.6cm} P{3.6cm} P{3.6cm}}
\toprule
\multirow[c]{3}{*}{Split} &
\multicolumn{4}{c}{Difficulty Group} \\
\cmidrule(lr){2-5}
 & Q1 -- Hard & Q2 -- Med-Hard & Q3 -- Med-Easy & Q4 -- Easy \\
\midrule
\midrule
Training (38) &
P017, P018, P022, P026, P034, P035, P042, P045, P050, P056 &
P001, P002, P003, P007, P012, P021, P033, P040, P048 &
P004, P014, P016, P032, P036, P046, P047, P052, P053, P054 &
P005, P010, P020, P028, P029, P037, P039, P041, P057 \\\hdashline
Validation (8) &
P038, P055 &
P009, P023 &
P006, P013 &
P019, P030 \\\hdashline
Testing (12) &
P008, P025, P044 &
P011, P024, P043 &
P015, P031, P058 &
P027, P051, P059 \\
\bottomrule
\end{tabular}
\begin{tablenotes}[para,flushleft]
\scriptsize
\item Q1: $z < -0.46$;\quad Q2: $-0.46 \leq z < -0.05$;\quad
Q3: $-0.05 \leq z < +0.40$;\quad Q4: $z \geq +0.40$.
\end{tablenotes}
\end{threeparttable}
\end{center}
\end{table*}

\subsection{Video}
\label{sec:video}
\begin{figure}
\begin{center}
\includegraphics[scale=0.56]{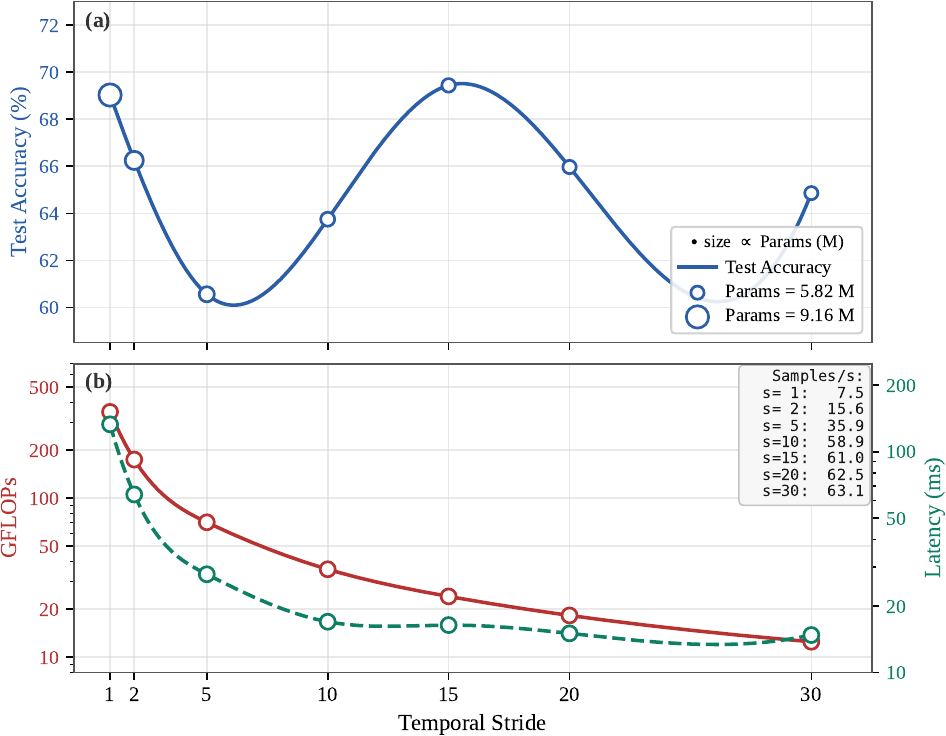}
\end{center}
\caption{Performance and computational cost of FUSE across temporal stride configurations.
(a)~Test accuracy and parameter count as a function of stride $\tau$; circle area encodes the number of parameters, with the smallest and largest sizes corresponding to $5.82$ M ($\tau=30$) and $9.16$ M ($\tau=1$), respectively.
(b)~GFLOPs (solid red, left axis) and inference latency in milliseconds (dashed green, right axis) as a function of stride, shown on logarithmic axes; throughput in samples per second is annotated for each configuration. Inference cost is measured on an NVIDIA A100 GPU.}

\label{performances}
\end{figure}

Table~\ref{table:videos} reports performance and computational cost across all seven temporal stride configurations; Fig.~\ref{performances} visualizes the accuracy and efficiency trends jointly. At the densest setting ($\tau=1$), the full $120$-second recording is retained at $30$~fps, yielding $L=3{,}600$ frames and a token channel dimension of $C'=10{,}826$. This configuration has the highest parameter count ($9.16$M), computational cost ($348.78$ GFLOPs), and inference latency ($133.02$ ms), with a corresponding throughput of $7.52$ samples per second, yet achieves a test accuracy of $69.03\%$.

Increasing the stride reduces $L$ proportionally, which contracts the token channel dimension $C' = 3L + 26$ and the associated input projection weights. At $\tau=30$, only $120$ frames are retained, reducing the parameter count to $5.82$M, GFLOPs to $12.48$, and latency to $14.77$ ms, corresponding to an approximately $28$-fold reduction in compute relative to $\tau=1$. As shown in Fig.~\ref{performances}(b), computational cost decreases sharply as the stride increases, while latency drops rapidly up to $\tau=10$ and then changes more gradually. Throughput similarly increases steeply at lower strides before saturating near $62$--$63$ samples per second for $\tau \geq 15$.

Test accuracy does not decrease monotonically with stride. The highest test accuracy is reached at $\tau=15$ ($69.44\%$), with $\tau=1$ yielding the second-best result ($69.03\%$). Validation accuracy peaks at $\tau=20$ ($70.42\%$). The configuration $\tau=5$ produces the lowest test accuracy ($60.56\%$) despite its second-highest validation accuracy ($68.80\%$), reflecting per-stride generalization variability over the $12$-subject test partition. The remaining configurations fall between $63.75\%$ and $66.25\%$ on the test set. These results indicate that temporal density beyond a moderate threshold does not yield consistent discriminative benefit, and that FUSE accommodates the full stride range without structural modification.

\begin{table*}
\caption{Performance and computational cost using the video modality.}
\label{table:videos}

\begin{center}
\begin{threeparttable}
\begin{tabular}{P{1.5cm} P{0.7cm} P{1.4cm} P{1.1cm}  P{2.5cm} P{2.0cm} P{0.9cm} P{0.9cm} P{0.4cm} P{1.0cm}}
\toprule

\multicolumn{2}{c}{Input} &
\multicolumn{2}{c}{Computational Cost} &
\multicolumn{2}{c}{Inference Cost} &
\multicolumn{3}{c}{Validation} &
\multicolumn{1}{c}{Testing} \\

\cmidrule(lr){1-2}\cmidrule(lr){3-4}\cmidrule(lr){5-6}\cmidrule(lr){7-9}\cmidrule(lr){10-10}
Modality &Stride & Params (M) & GFLOPs &Latency (ms) GPU$\downarrow$  &Samples/s GPU$\uparrow$ & Accuracy & Precision & F1 &Accuracy  \\

\midrule
\midrule
Video & 1  &9.16 &348.78 &133.02 &7.52 &63.90 &66.70 &60.48 &69.03 \\\hdashline
Video & 2  &7.43 &174.83 &64.00 &15.63 &66.42 &67.13 &66.52 &66.25 \\\hdashline
Video & 5  &6.39 &70.46  &27.84 &35.92 &68.80 &69.39 &67.39 &60.56 \\\hdashline
Video & 10 &6.05 &35.67  &16.97 &58.92 &66.39 &66.20 &66.14 &63.75 \\\hdashline
Video & 15 &5.93 &24.07  &16.39 &61.01 &67.77 &68.33 &67.89 &69.44 \\\hdashline
Video & 20 &5.87 &18.27  &15.03 &62.47 &70.42 &71.86 &68.45 &65.97 \\\hdashline
Video & 30 &5.82 &12.48  &14.77 &63.14 &68.60 &69.91 &68.72 &64.86
\\

\bottomrule
\end{tabular}
\begin{tablenotes}[para,flushleft]
\scriptsize
\item  Inference Cost measured on an NVIDIA A100 GPU.
\end{tablenotes}
\end{threeparttable}
\end{center}
\end{table*}

\subsection{Overall Analysis \& Discussion}

The central capability demonstrated by this evaluation is that FUSE processes the complete facial recording as a single input, without temporal windowing or external segmentation. At $\tau=1$, this corresponds to $3{,}600$ frames from a $120$-second recording ingested in one model pass. This differs from common video-based stress-recognition pipelines that first decompose recordings into short clips or predefined temporal windows before classification. In FUSE, the full sequence is encoded via axis folding and processed with asymmetric attention, whereas the internal spatial token segmentation does not correspond to temporal windows.
The reorganization of the temporal axis is consistent with prior evidence that restructuring facial spatiotemporal representations can benefit affective assessment \cite{gkikas_reface_acii_2026}.

The stride sweep shows that this capability is not limited to dense sampling. The same architecture accommodates inputs ranging from $120$ frames at $\tau=30$ to $3{,}600$ frames at $\tau=1$, with the token channel dimension $C' = 3L + 26$ scaling with the retained sequence length. No clip-level aggregation, temporal pooling, or external segmentation is introduced. This supports the main design goal of FUSE: any-length facial video processing without changing the model structure across temporal-stride settings.

The best test accuracy is obtained at $\tau=15$ ($69.44\%$), using $24.07$ GFLOPs, compared with $348.78$ GFLOPs at $\tau=1$. The full-frame configuration still achieves a competitive test accuracy of $69.03\%$ while processing the entire $3{,}600$-frame sequence. The non-monotonic relationship between stride and test accuracy, visible in Fig.~\ref{performances}(a), indicates that denser temporal sampling does not necessarily improve generalization. Instead, moderate subsampling can preserve stress-relevant facial information while substantially reducing computational cost.
The higher strides likely perform comparably because consecutive frames at $30$~fps are largely redundant, so retaining every frame mainly enlarges the input projection ($5.82$M to $9.16$M parameters) without adding useful information.

A limitation of the current evaluation is that it uses a single dataset and a binary neutral-versus-stress classification setting. In addition, the study does not include a direct comparison against windowed baselines under the same subject-level split, which is left for future work. Windowing itself offers a practical benefit, as fixed-length clips yield a smaller and uniform input that avoids the growing channel dimension of the full-recording formulation. Consequently, the results support the feasibility of full-recording inference, but they should not be interpreted as a complete replacement for all window-based video stress-recognition strategies.

\section{Conclusion}
This paper presented FUSE, a facial-video stress detection framework designed to process complete recordings without temporal windowing or external segmentation. The proposed approach folds the temporal dimension into the channel dimension and uses asymmetric attention to process the resulting high-dimensional representation through a compact set of segment states. This enables the same architecture to operate across a wide range of temporal stride settings, from sparse subsampling to full-frame input.
Experiments on a $58$-subject stress dataset show that FUSE can process a full $120$-second recording at $\tau=1$, corresponding to $3{,}600$ frames, while also supporting more efficient stride configurations without structural changes. The best test accuracy is achieved at $\tau=15$ ($69.44\%$), whereas the full-frame configuration remains competitive at $69.03\%$. These results indicate that temporal windowing is not required for effective facial-video stress detection in this setting, and that complete-recording inference can be achieved within a single unified architecture.
Overall, FUSE shifts the unit of analysis from short temporal clips to complete facial recordings. This provides a simpler evaluation pipeline, avoids choices about window length and aggregation, and preserves the ability to model stress-related facial information over the full duration of the recording.

\section*{Acknowledgments}
The authors used large language model (LLM)-based tools for language editing and improvement. All scientific content, results, and conclusions are solely the work of the authors.


\bibliographystyle{IEEEtran}
\bibliography{library}

\end{document}